# A Comparative Study of Removal Noise from Remote Sensing Image

Mr. Salem Saleh Al-amri[1], Dr. N.V. Kalyankar[2] and Dr. Khamitkar S.D [3]

[1] Research Student,
Computer Science Dept.,
Yeshwant College, Nanded

[2] Principal,
Yeshwant College, Nanded

[3]Director
School of Computational Science,
Swami Ramanand Teerth Marathwada University, Nanded, India)

## Abstract
This paper attempts to undertake the study of three types of noise such as Salt and Pepper (SPN), Random variation Impulse Noise (RVIN), Speckle (SPKN). Different noise densities have been removed between 10% to 60% by using five types of filters as Mean Filter (MF), Adaptive Wiener Filter (AWF), Gaussian Filter (GF), Standard Median Filter (SMF) and Adaptive Median Filter (AMF). The same is applied to the Saturn remote sensing image and they are compared with one another. The comparative study is conducted with the help of Mean Square Errors (MSE) and Peak-Signal to Noise Ratio (PSNR). So as to choose the base method for removal of noise from remote sensing image.

**Keywords:** Remote sensing image, Image noise, filters

**Introduction:**
Digital image processing is the most important technique used in remote sensing. It has helped in the access to technical data in digital and multi-wavelength, services of computers in terms of speed of processing the data and the possibilities of big storage. Several studies can also take the benefit of it such as technical diversity of the digital image processing, replication sites and maintaining the accuracy of the original data. Noise is removable using iterative median filtering in spatial domain which requires much less processing time than removal by frequency domain Fourier transforms [1]. Weight Median Filter (WMF) based on threshold decomposition removes impulsive noise with an excellent image detail processing capability compared to nonlinear filter and linear filter [2]. Standard Median Filtering (SMF) is a non-linear, low-pass filtering method which can be used to remove 'speckle' noise from an image. A median filter can out perform linear, low pass filters, on this type of noisy image became it can potentially remove all the noise without affecting the 'clean' pixels. Median filters remove isolated pixels, whether they are bright or dark. Adaptive Median Filter (AMF) is designed to eliminate the problems faced by the Standard Median Filter [3]. Adaptive Filter (AF) changes its behavior based on the statistical characteristics of the image inside the filter window. Adaptive filter performance is usually superior to non-adaptive counterparts. The improved performance is at the cost of added filter complexity. Mean and variance are two important statistical measures using which adaptive filters can be designed [4].There are many methods for reducing noise. Traditional median filter and mean filter are used to reduce salt-pepper noise and Gaussian noise respectively. When these two noises exist in the image at the same time, use of only one filter method can not achieve the designed result [5].

## 1. Remote Sensing Image
Remote sensing is used to obtain information about a target or an area or a phenomenon through the analysis of certain information which is obtained by the remote sensor. It does not touch these objects to verify. Images obtained by satellites are useful in many environmental applications such as tracking of earth resources, geographical mapping, prediction of agricultural crops, urban growth, weather, flood and fire control etc. Space image application includes recognition and analysis of objects in the images, obtained from deep space-probe missions.

## 2. Image Noise
Noise is any undesired information that contaminates an image. Noise appears in image from various sources. The digital image acquisition process, which converts an optical image into a continuous electrical signal that is then sampled, is primary process by which noise appears in digital image. There are several ways through which noise can be introduced into an image, depending on how the image is created.

Satellite image, containing the noise signals and lead to a distorted image and not being able to understand and study it properly, requires the use of appropriate filters to limit or reduce much of the noise. It helps the possibility of better interpretation of the content of the image.

### 2.1 Types of Noise
There are three common types of image nose:

### 2.1.1 Random Variation Impulsive Noise (RVIN)
This type of noise is also called the Gaussian noise or normal noise is randomly occurs as white intensity values. Gaussian distribution noise can be expressed by:

$$P(x) = 1/(\sigma\sqrt{2}\pi) * e^{(x-\mu)2}/2\sigma^2 \qquad -\infty < 0 < \infty \quad (1)$$

Where: $P(x)$ is the Gaussian distribution noise in image; $\mu$ and $\sigma$ is the mean and standard deviation respectively.

### 2.1.2 Salt & Pepper Noise (SPN)
This type contains random occurrences of both black and white intensity values, and often caused by threshold of noise image.
Salt & pepper distribution noise can be expressed by:

$$p(x) = \begin{cases} p1, & x = A \\ p2, & x = B \\ 0, & otherwise \end{cases} \quad (2)$$

Where: $p1$, $p2$ are the Probabilities Density Function (PDF), $p(x)$ is distribution salt and pepper noise in image and $A$, $B$ are the arrays size image. Gaussian and salt & Pepper are called impulsive noise.

### 2.1.3 Speckle Noise (SPKN)
If the multiplicative noise is added in the image, speckle noise is a ubiquitous artifact that limits the interpretation of optical coherence of remote sensing image. The distribution noise can be expressed by:

$$J = I + n*I \quad (3)$$

Where, $J$ is the distribution speckle noise image, $I$ is the input image and $n$ is the uniform noise image by mean $o$ and variance $v$.

## 3. Filters
### 3.1 Mean Filter (MF)
Mean Filter (MF) is a simple linear filter, intuitive and easy to implement method of smoothing images, i.e. reducing the amount of intensity variation between one pixel and the next. It is often used to reduce noise in images. The idea of mean filtering is simply to replace each pixel value in an image with the mean (average) value of its neighbors, including itself. This has the effect of eliminating pixel values which are unrepresentative of their surroundings. Mean filtering is usually thought of as a convolution filter. Like other convolutions it is based around a kernel, which represents the shape and size of the neighborhood to be sampled when calculating the mean.

### 3.2 Standard Median Filter (SMF)
Median filter is the non-linear filter which changes the image intensity mean value if the spatial noise distribution in the image is not symmetrical within the window. Median filter reduce is the variance of the intensities in the image. Median filter is a spatial filtering operation, so it uses a 2-D mask that is applied to each pixel in the input image. To apply the mask means to centre it in a pixel, evaluating the covered pixel brightness and determining which brightness value is the median value.

### 3.3 Adaptive Wiener Filter (AWF)
Adaptive Wiener Filter (AWF) changes its behavior based on the statistical characteristics of the image inside the filter window. Adaptive filter performance is usually superior to non-adaptive counterparts. But the improved performance is at the cost of added filter complexity. Mean and variance are two important statistical measures using which adaptive filters can be designed.

### 3.4 Gaussian Filter (GF)
Gaussian low pass filter is the filter which is impulse responsive, Gaussian filters are designed to give no overshoot to a step function input while minimizing the rise and fall time. Gaussian is smoothing filter in the 2D convolution operation that is used to remove noise and blur from image.

### 3.5 Adaptive Median Filter (AMF)
The Adaptive Median Filter (AMF) is designed to eliminate the problems faced with the Standard Median Filter. The basic difference between the two filters is that in the Adaptive Median Filter, the size of the window surrounding each pixel is variable. This variation depends on the median of the pixels in the present window. If the median value is an impulse, then the size of the window is expanded.

## 4. Experiments Verifications
### 4.1 Testing Procedure
The filters were implemented using (MATLAB R2007a, 7.4a) and tested three types of noise: Speckle Noise (SPKN), Random Variation Impulsive Noise (RVIN) and Salt & Pepper Noise (SPN) corrupted on the Saturn image illustrated in the Fig. 1.

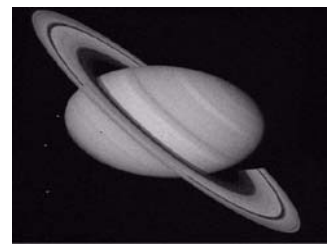

**Fig.1 – Saturn Image**

For this image, its performance for (SPN), (RVIN) and (SPKN), noise with probabilities from 10% to 60%.
Five types of filters are implemented. Mean Filter, Median Filter, Adaptive Filter, LPF Gaussian Filter, Adaptive Median Filter.

### 4.2 Simulation Results
Intensive simulations were carried out using one monochrome satellite images are chosen for demonstration. The performance evaluation of the filtering operation is quantified by the PSNR (Peak Signal to Noise Ratio) and MSE (Mean Square Error) calculated using formula:

$$PSNR = 10 log_{10}\left(\frac{255^2}{MSE}\right)$$

Where,

$$MSE = \frac{1}{MN} \sum_{i=1}^{M} \sum_{i=1}^{N} [g(i,j) - f(i,j)]^2$$

Where, $M$ and $N$ are the total number of pixels in the horizontal and the vertical dimensions of image. $g$ denotes the Noise image and $f$ denotes the filtered image.

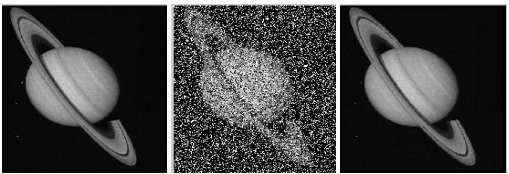

**Fig.2– Image corrupted 60% (SPN) and filtered by Adaptive Median Filter (AMF)**

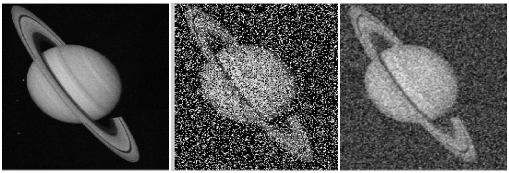

(a) Original     (b) 60%Noisily     (c) AWF

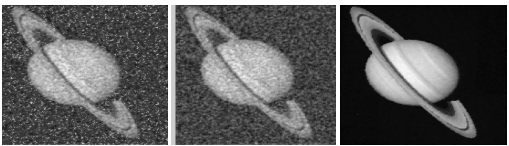

(d) GF     (e) MF     (f) SMF

**Fig.3 – Image corrupted by 60% salt & Pepper Noise**

**Table 1: Restoration Result PSNR for Salt and Pepper Noise**

| Filter Type | 10% | 20% | 30% | 40% | 50% | 60% |
|---|---|---|---|---|---|---|
| MF | 33.65 | 31.92 | 30.77 | 29.90 | 29.23 | 28.65 |
| AWF | 33.74 | 31.93 | 30.78 | 29.90 | 29.23 | 28.65 |
| GF | 33.78 | 31.95 | 30.79 | 29.91 | 29.24 | 28.67 |
| SMF | 34.30 | 32.58 | 31.34 | 30.40 | 29.62 | 28.95 |

**Table 2: Restoration Result MSE for Salt & Pepper Noise**

| Filter Type | 10% | 20% | 30% | 40% | 50% | 60% |
|---|---|---|---|---|---|---|
| MF | 28.07 | 41.82 | 54.51 | 66.54 | 77.65 | 88.64 |
| AWF | 27.52 | 41.71 | 54.39 | 66.53 | 77.65 | 88.64 |
| GF | 27.24 | 41.50 | 54.24 | 66.38 | 77.55 | 88.39 |
| SMF | 24.19 | 35.86 | 47.82 | 59.32 | 71.01 | 82.74 |

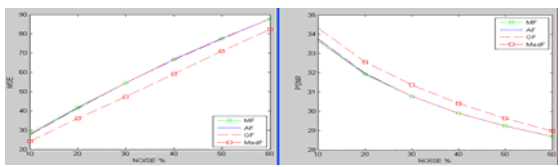

**Fig.4 – PSNR and MSE Analyses for (SPN)**

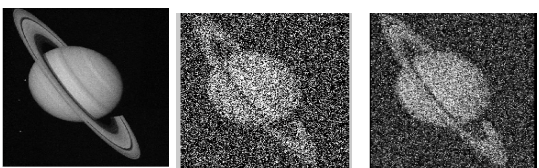

(a) Original     (b) 60 %Noisily     (c) AMF

**Fig.5 – Image corrupted by 60% (RVIN) Noise and Filtered by (AMF)**

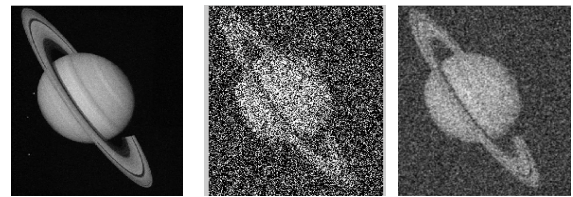

(a) Original     (b) 60 %Noisily     (c) AWF

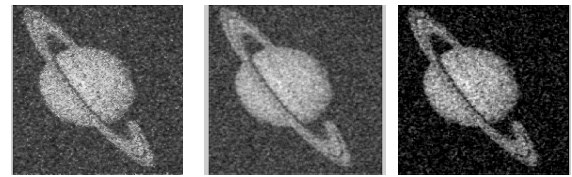

(d) GF     (e) MF     (f) SMF

**Fig.6– Image corrupted by 60% (RVIN)**

**Table 3: Restoration Result PSNR for (RVIN)**

| Filter Type | 10% | 20% | 30% | 40% | 50% | 60% |
|---|---|---|---|---|---|---|
| MF | 28.53 | 28.40 | 28.30 | 28.23 | 28.19 | 28.14 |
| AWF | 28.63 | 28.45 | 28.34 | 28.25 | 28.21 | 28.15 |
| GF | 28.56 | 28.42 | 28.32 | 28.25 | 28.21 | 28.15 |
| SMF | 27.68 | 27.56 | 27.51 | 27.48 | 27.46 | 27.45 |

| Filter Type | 10% | 20% | 30% | 40% | 50% | 60% |
|---|---|---|---|---|---|---|
| MF | 91.24 | 94.04 | 96.11 | 97.81 | 98.60 | 99.88 |
| AWF | 89.17 | 92.93 | 95.41 | 92.32 | 98.21 | 99.56 |
| GF | 90.50 | 93.56 | 95.65 | 97.33 | 98.18 | 99.55 |
| SMF | 110.94 | 114.14 | 115.46 | 116.26 | 116.68 | 116.9 |

**Table 4: Restoration Result MSE for Salt and (RVIN)**

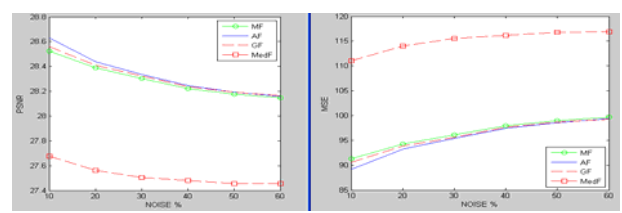

**Fig. 7 – PSNR and MSE Analyses for (RVIN) Noise**

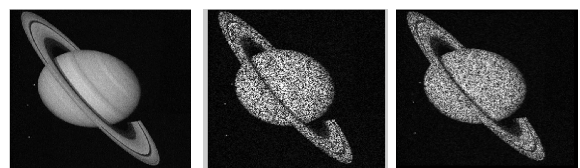

(a) Original     (b) 60 % Noisily     (c) AMF

**Fig.8 – Image Corrupted by 60% Speckle Noise**

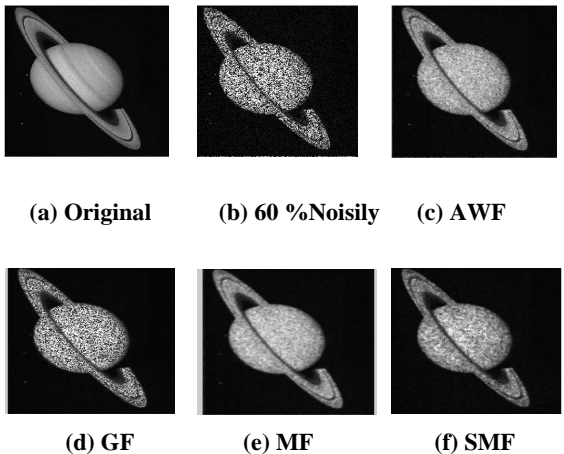

**(a) Original**  **(b) 60 %Noisily**  **(c) AWF**

**(d) GF**  **(e) MF**  **(f) SMF**

**Fig.9 – Image Corrupted by 60% Speckle Noise**

**Table 5: Restoration Result PSNR for Speckle Noise**

| Filter Type | 10% | 20% | 30% | 40% | 50% | 60% |
|---|---|---|---|---|---|---|
| MF | 32.21 | 31.62 | 31.23 | 30.92 | 30.66 | 30.50 |
| AWF | 33.54 | 32.47 | 31.86 | 31.43 | 31.10 | 30.89 |
| GF | 32.40 | 31.80 | 31.40 | 31.08 | 30.83 | 30.65 |
| SMF | 31.98 | 31.37 | 30.99 | 30.68 | 30.43 | 30.24 |

**Table 6: Restoration Result MSE for Speckle Noise**

| Filter Type | 10% | 20% | 30% | 40% | 50% | 60% |
|---|---|---|---|---|---|---|
| MF | 39.12 | 44.79 | 48.94 | 52.61 | 55.81 | 57.91 |
| AWF | 28.79 | 36.82 | 42.39 | 46.78 | 50.50 | 53.04 |
| GF | 37.41 | 43.00 | 47.08 | 50.67 | 53.72 | 55.93 |
| SMF | 41.25 | 47.40 | 51.76 | 55.57 | 58.84 | 61.57 |

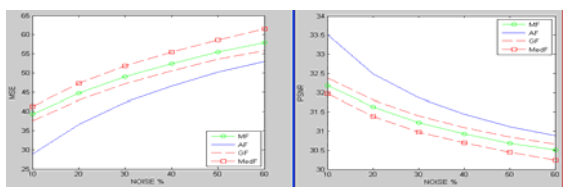

**Fig.10 – PSNR and MSE Analyses for 60% Speckle Noise**

## 5. Conclusion

In this paper, the comparative studies are explained & experiments are carried out for different filters, Adaptive Median Filter (AMF) is the best filter to remove SPN noise of image sensing and see this in the figure(2). It doesn't leave any blurring in the image and Standard Median Filter (SMF) is good filter for SPN with less than 40% density noise see in the figure(3). The comparative study explained with help of PSNR and MSE which illustrated in the figures (4), (7) and (10) with values illustrated in the table (1), (2),(4),(5) and (6). The best results of Adaptive Filter (AWF), Gaussian Filter (GF), Main Filter (MF) and Standard Median Filter (SMF) respectively with small difference between them we also can use Adaptive Median Filter (AMF) of the small density noise, but in RVIN noise the comparative study explains the best result of MF, AWF, and GF respectively in the small density noise with fails of the SMF and AMF in this type of noise see this in the figures (5) and (6). In SPKN noise the best of the MF and AWF respectively with not bad for other filters see in the figure (8) and (9).

## AUTHORS

**Dr.N.V. Kalyankar**. BSc.Maths, Physics, Chemistry, Marathwada University, Aurangabad, India, 1978. M Sc.Nuclear Physics, Marathwada University, AurangabadIndia, 1980.Diploma in Higher Education, Shivaji University, Kolhapur, India, 1984.Ph.D. in Physics, Dr. B.A.M.University, Aurangabad, India, 1995.Principal Yeshwant Mahavidyalaya College, Membership of Academic Bodies, Chairman, Information Technology Society State Level Organization, Life Member of Indian Laser Association, Member Indian Institute of Public Administration, New Delhi, Member Chinmay Education Society, Nanded.he has one publication book, seven journals papers, two seminars Papers and three conferences papers.

**Dr. S. D. Khamitkar.** MSc. Ph.D. Computer Science Reader & Director (School of Computational Science) Swami Ramanand Teerth Marathwada University, Nanded,14 Years PG Teaching, Publications 08 International, Research Guide (10 Students registered),Member Board of Studies (Computer Application),Member Research and Recognition Committee (RRC) (Computer Studies).

**Mr. Salem Saleh Al-amri.** Received the B.E degree in, Mechanical Engineering from University of Aden, Yemen, Aden in 1991, the M.Sc.degree in, Computer science (IT) from North Mahrashtra University(N.M.U), India, Jalgaon in 2006, Research student Ph.D in the department of computer science (S.R.T.M.U), India, Nanded.

### Acknowledgements

We place our heart-felt gratitude to Dr.N.V.Kalyankar,Principal,Yeshwant College, Nanded and Dr. Khamitkar S.D , Derector of School of Computational Science, Swami Ramanand Teerth Marathwada University, Nanded.for them most valuable guidance without which this work would have been impossible.

### REFERENCE

[1] Nichol, J.E. and Vohra, V., Noise over water surfaces In Landsat TM images, International Journal of Remote Sensing, Vol.25, No.11, 2004, PP.2087 - 2093.

[2] Mr. F. N. Hasoon, Weight Median Filter Using Neural Network for Reducing Impulse Noise,M.S.thesis,Department Computer Sciences, University of Putra,Putra, Malaysia,2008.

[3] D.Dhanasekaran, K. Bagan, High Speed Pipeline Architecture for Adaptive Median Filter, European Journal of Scientific Research, Vol.29, No.4, 2009, PP.454-460.

[4] R.C.Gonzalez and R.E. Wood, Digital Image Processing, Prentice-Hall, India, Second Edition, 2007.

[5] Chi Chang-Yanab, Zhang Ji-Xiana, Liu Zheng-Juna, Study on Methods of Noise Reduction in a Stripped Image, the International Archives of the Photogrammetry, Remote Sensing and Spatial Information Sciences. Vol XXXVII. Part B6b, Beijing, 2008.

[6] K.Amolins, Y.Zhang, P.Dare. Wavelet-based image fusion techniques- An introduction, review and camprarison, ISPRS Journal of photogrammetry Remote Sensing, Vol.62, No.4, 2007, PP.249-263.

[7] A.K.Jain, Fundamentals of Digital Image Processing, University of California, Davis, Prentice Hall of India Private Limited, New Delhi-110001, 2007.

[8] R .H.Chan ,W.H.Chung and M.Nikolova, Salt-and-Pepper Noise Removal by Median-Type Noise Detectors and Details-Preserving Regularization, Image processing,IEEE,Vol.14,No.10,2005,PP.1479-1485.

[9] E.S. Gopi, Digital Image Processing using Matlab, Senior Lecturer, Department of Electronics and Communication Engineering, Sri. Venkateswara College of Engineering Pennalur, Sriperumbudur, Tamilnadu, SciTech Publication (India) Pvt. Ltd., 2007.

[10] M.S.Alani, Digital Image Processing using Matlab, University Bookshop, Sharqa, URA, 2008.

[11] M.A.Joshi, Digital Image Processing – An Algorithmic Approach, Professors and Head, Department of Telecommunications, College of Engineering, Pune,Prentice Hall of India Private Limited, New Delhi, 2007.